%% file: example_paper.tex
\documentclass{article}

\usepackage{microtype}
\usepackage{graphicx}
\usepackage{booktabs} 

\usepackage{hyperref}

\usepackage[accepted]{icml2018}

\graphicspath{{./figures/}}
\usepackage{amsfonts}
\usepackage{amsmath}
\usepackage{amsthm}
\usepackage{mathtools}
\usepackage{subcaption}
\usepackage{bbm}


\icmltitlerunning{Deep Structured Generative Models}

\begin{document}

\twocolumn[
\icmltitle{Deep Structured Generative Models}



\icmlsetsymbol{equal}{*}

\begin{icmlauthorlist}
\icmlauthor{Kun Xu}{equal,dcs}
\icmlauthor{Haoyu Liang}{equal,dcs}
\icmlauthor{Jun Zhu}{dcs}
\icmlauthor{Hang Su}{dcs}
\icmlauthor{Bo Zhang}{dcs}
\end{icmlauthorlist}

\icmlaffiliation{dcs}{
Dept. of Comp. Sci. \& Tech., BNRist Center, State Key Lab for Intell. Tech. \& Sys., THBI Lab, Tsinghua University, Beijing, 100084, China
}
\icmlcorrespondingauthor{Jun Zhu}{dcsjz@mail.tsinghua.edu.cn}

\icmlkeywords{Deep Learning, Deep Generative Models}

\vskip 0.3in
]



\printAffiliationsAndNotice{\icmlEqualContribution} 

\begin{abstract}
Deep generative models have shown promising results in generating realistic images, but it is still non-trivial to generate images with complicated structures. The main reason is that most of the current generative models fail to explore the structures in the images including spatial layout and semantic relations between objects.
To address this issue, we propose a novel deep structured generative model which boosts generative adversarial networks (GANs) with the aid of structure information. In particular, the layout or structure of the scene is encoded by a stochastic and-or graph (sAOG), in which the terminal nodes represent single objects and edges represent relations between objects. 
With the sAOG appropriately harnessed, our model can successfully capture the intrinsic structure in the scenes and generate images of complicated scenes accordingly. Furthermore, a detection network is introduced to infer scene structures from a image. Experimental results demonstrate the effectiveness of our proposed method on both modeling the intrinsic structures, and generating realistic images.
\end{abstract}

\section{Introduction}

Deep generative models (DGMs)~\cite{kingma2013auto,goodfellow2014generative,li2015generative} have made a great progress in image generation. 
However, realistic images have complicated intrinsic structures and most DGMs fail to consider these structures by simply mapping a vector of random noise into image space. 
By incorporating valid structures explicitly, images with complicated relations can be generated conditionally on these complicated structures. Besides, with these highly-interpretable structures, a better understanding of scene generation can be got, and a set of operations, such as moving objects, changing locations, become possible. Indeed, ~\citet{mirza2014conditional} and~\citet{chongxuan2017triple} introduce conditional variables to generative models as an input, and synthesize images conditionally.
But using a one-hot conditional variable is far from modeling the relations within a complicated realistic image.

Image grammar models~\cite{zhu2007stochastic} provide a principled way to model the structure of images, which have been widely used in modeling and parsing the hierarchical scene structures from images~\cite{zhao2011image}.
The models define a hierarchical structure of the whole scenes explicitly, which consequently facilitate an explicit modeling of the object relations and an embedding of human-knowledge for certain configurations. Generally, 
a grammar model starts from a node representing the complicated scene, and ends with nodes, each of which represents a single object. 
Production rules are used to define the composition of the visual elements hierarchically, and the relations between objects are introduced by edges between nodes.
With a probability defined on graphs, grammar models can successfully capture the structure information of the complicated scene, as well as the uncertainty.

However, grammar models struggle with the expressive ability, i.e., it can hardly model complicated appearance for each object. The inability mainly arises from the fact that grammar models use very low level visual primitives, such as lines~\cite{liu2014single}, to present complicated images. 
Recently, the great advances made by DGMs provides an option for bridging the gap between the structure model and realistic images. Especially with the emergence of Generative Adversarial Networks (GANs)~\cite{goodfellow2014generative}, images with high-resolutions, (e.g., 1024$\times$1024) can be generated successfully~\cite{karras2017progressive}.  

\begin{figure}[hbt]
	\centering
	\includegraphics[width=0.45\textwidth]{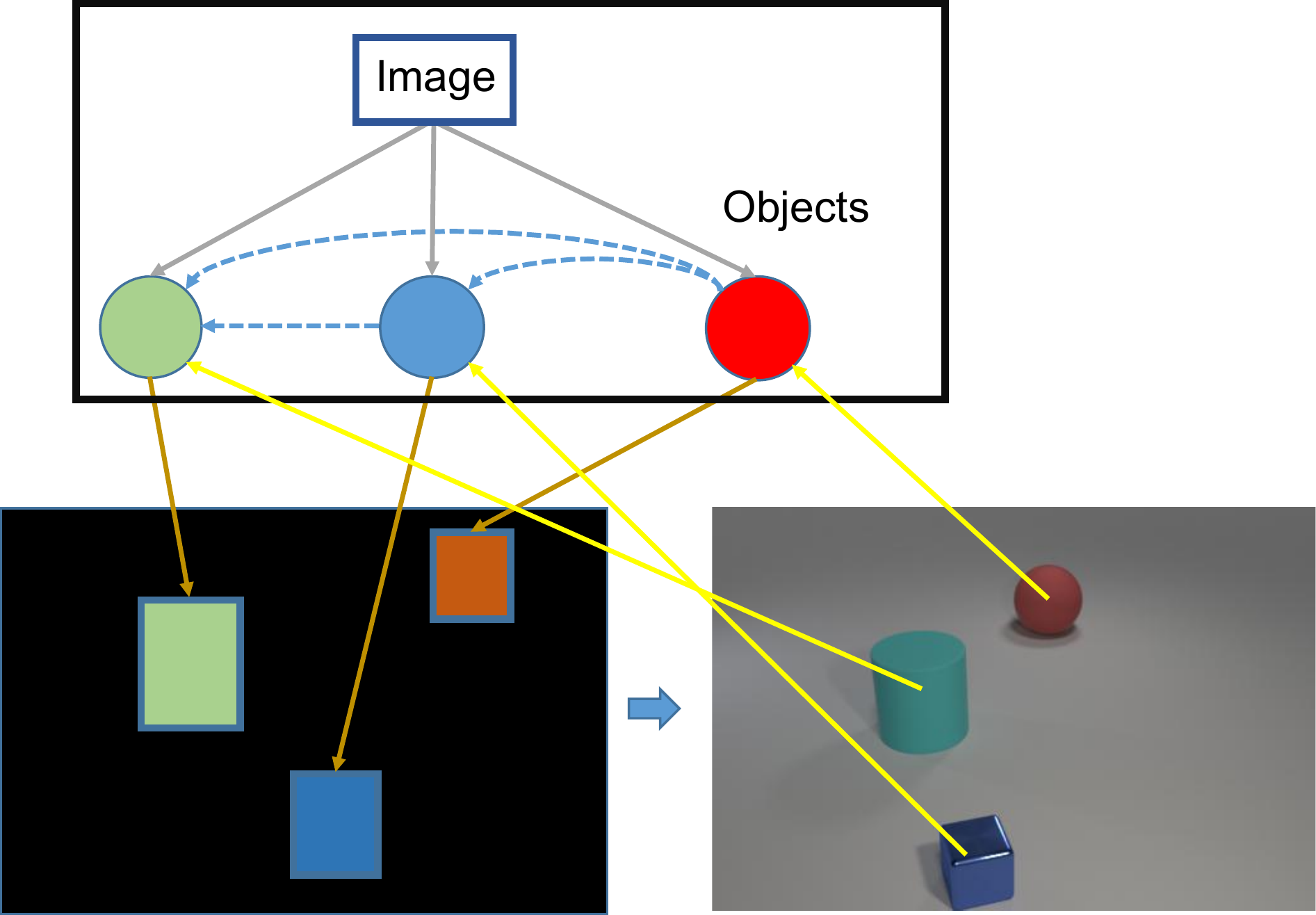}
	\caption{A simple demo of our proposed Method. A structured structure variable (black box) is introduced to model the relations between objects. With the relations (dashed arrow) explicit modeled, a layout configuration (left) is generated by a grammar model, and then a deep generative model refines it into realistic images (blue arrow). Besides, a recognition model (yellow arrows) is deployed in this framework to infer the structure given an image.}
	\label{fig:demo}
\end{figure}

\subsection{Our proposal}

To incorporate structure models in image generation, we propose a novel framework for image generation, which can conjoin the respective advantages of deep generative models and grammar models. Specifically, a layout configuration is modeled using a grammar model, which can model a wide variety of intrinsic structures explicitly. 
A generative model is used to bridge the gap between structure informations and visual primitives by generating textures, colors and other primitives according to the configuration. Besides, we also apply a detection model to infer the latent structure from a image by finding the most possible valid configuration for the current image.

Fig.~\ref{fig:demo} presents an illustration of our method.
The proposed framework is a hierarchical variable model, where rather than using a vector of random noise, a stochastic grammar model is used as the structure variable to capture structure information.

Specifically, the Stochastic And-Or Graph (sAOG)~\cite{zhu2007stochastic} is used as an image grammar to model the 3D arrangement and relations between objects, which is further boosted by taking the camera location into account as a prior to generate a bounding box of each object for further generation.
Then we construct a pixel-to-pixel refinement network, conditioning on the bounding box generated by the image grammar, and attributes of objects. Different from the previous methods~\cite{zhao2011image,tu2002image}, it is not necessary for the sAOG to generate pixelwise images but to produce the high-level layout of the scenes, which can be fed to the deep generative models as conditions to guide the generation of realistic images.

More specifically, we introduce a Stochastic And-Or Graph (sAOG) to model the structure of scene images. In sAOG, a scene is represented as a parse tree, where it starts with a single root node (the scene) and ends with a set of terminal nodes (single objects). Nonterminal nodes represent different levels of structures or groups. sAOG successfully models the hierarchical structure of a visual instance, where the relations between objects and groups can be modeled by the edge between nodes. 
A sample from sAOG is called $parse$ $graph$ which represents a scene configuration with certain set of objects and relations.
sAOG also provides a good opportunity to embed human knowledge of scene layout into the model structures, e.g., we can ensure locations satisfy the constrains according to their spatial or semantic relations. The parameters can be learned from the dataset~\cite{zhu1997minimax} by maximizing likelihood estimation.

With the sAOG embedded, we adopt a generator called refinement network parameterized as pixel-to-pixel model~\cite{isola2017image} to refine the coarse bounding box images to realistic images. It enables us to integrally generate a scene image with rich appearance conditioned on a map of attribution, such as spatial position, orientation, material, color, etc. The adopted deep generative model alleviates the shortage of limited expressive power of grammar models.

Besides generation, a recognition model is proposed to infer the latent structure from a scene image. In this work, we apply an object detection model to provide bounding boxes of objects as well as corresponding attributes.
After that, a submodule takes the visual features and coordinates of each bounding box as input to regress 3D coordinates.
With objects detected, a sAOG with corresponding object nodes is created, and relations between them are inferred by maximizing a posterior using markov chain monte carlo (MCMC) method.

Overall, our contributions can be summarized into three-fold: 
\begin{itemize}   
	\item To our best knowledge, this is the first attempt to integrating the stochastic grammar technique and deep generative models within a unified framework, such that the images with complex relations can be generated effectively and conditionally with the scene configurations appropriately harnessed;
    \item We propose a novel framework to characterize the layout of image explicitly such that the relation among the objects can be modeled in a flexible way. It therefore not only alleviates the difficulties of training GAN for large-scale images, but facilitates a deeper understanding of the complicated scene configuration.  
    \item An inference network is introduced to recognize the objects to help grammar model infer the parse graph given a complicated image. It not only alleviates the shortage of expressive ability in recognition part, but boosts the speed of MCMC by shrinking the search space from image space to the space of 3D locations and relations.
	
\end{itemize}

\section{Backgrounds}
In this section, we first give a brief introduction to stochastic and-or graph, then deep generative models and pixel-to-pixel translation are introduced.

\subsection{Stochastic Image Grammar}

Stochastic image grammar~\cite{zhu2007stochastic} is a probabilistic framework to model image structures, usually formulated as stochastic And-Or Graph (sAOG)\cite{zhu2007stochastic}. sAOG is defined as a 5-tuple $<S, \boldsymbol{V}, \boldsymbol{R},\boldsymbol{E}, P>$, where 1) $S$ denotes a root node for the whole scene; 2) $\boldsymbol{V}$ is a finite vertex set containing terminal node set $\boldsymbol{V}_T$ representing the primitive units in a scene like objects, and non-terminal node set $\boldsymbol{V}_N$ representing higher level structures; 3) $\boldsymbol{R}$ is a set of production rules where each decomposes a specific node in $\boldsymbol{V}_N$ to a certain sequence of child node(s) in $\boldsymbol{V}$; 4) $\boldsymbol{E} \subset  \boldsymbol{V}\times \boldsymbol{V}$ is a set of relations between nodes where each type of relation can be represented as a type of edge in and-or graph; 5) We call a graph generated by this grammar as a parse graph. $P$ is a probability function that defines the probability for all valid parse graph.


Stochastic image grammar can be applied on both discriminative approaches~\cite{song2013discriminatively} and  generative approaches~\cite{zhao2011image,Qi2018human}. In \cite{park2015attributed,fang2018learning}, the authors define a grammar on the human pose, and the 3D pose and parts can be inferred from the grammar.
\citet{liu2014single} define an attribute grammar on the man-made world for parsing and recover the 3D spatial information from single-view images. In this grammar, nodes denote surfaces  in space or combination of them with their geometric attributes.
Most previous methods on sAOG use low-level elements as the terminal nodes, such as line segment, rectangles and image patch, which limits performance of the grammar.


Another important structured image generation approach is to use scene graph as condition~\cite{johnson2018image}. A scene graph is represented by a set of nodes for objects and a set edges for relations. 
In this work, a scene graph is first converted to image space using a graph convolutional network, and then they apply an image-to-image model to refine the images. However, there's no hierarchical structures within the scene graph, and
neither modeling uncertainties nor generating scene graphs can be achieved because the scene graph is generally deterministic rather than stochastic.

\subsection{Deep generative models}

Deep generative models have made great advances in image generation recently, among which Generative Adversarial Nets (GANs)~\cite{goodfellow2014generative} show a powerful ability to implicitly model the distribution on image data and generate realistic images.
In GANs, a generator (G) transfers a input noise into data manifold trying to fool the discriminator, and a discriminator (D) tries to distinguish whether a sample is from the data distribution, which provides training signal for generator. 
Thus objective function can be formulated under a minimax game framework:
\begin{eqnarray}
\min_G \max_D V(G,D) = \mathbb{E}_{x\sim p_{ \mathtt{data}} }[\log D(X;\Phi)]\nonumber\\ 
+\mathbb{E}_{z\sim p(z)}[\log(1-D(G(z;\theta);\Phi))], \nonumber
\end{eqnarray}
where the $p(z)$ generally is a standard normal distribution as the noise input and both G and D are neural networks parameterized by $\theta$ and $\Phi$ respectively. Both G and D can be trained using back-propagation. Under this framework, the generator can recover the data distribution when global equilibrium is achieved.


To generate complicated images, image-to-image translation based on GANs are proposed to generate image conditionally. In S$^2$-GAN ~\cite{wang2016generative}, an image-to-image adversarial network takes a  map of surface normal as condition and outputs another realistic image, where the normal map is generated by another vanilla GAN to represent the structure of a scene. 
By using an  U-net generator and a convolutional discriminator, pix2pix model~\cite{isola2017image} performs well on a wider variety of  image-to-image tasks, including style transferring and  reconstructing images from label maps or edge maps.
Pix2pixHD~\cite{wang2017high} introduces multi-scale discriminators and an instance encoder to pix2pix, so as to generate high resolution realistic images conditioned on instance  segmentation maps. Besides that,  pix2pixHD enables users to edit image structures by manipulating instance maps.

\section{Method}
\begin{figure*}
	\centering
	\includegraphics[width=0.95\textwidth]{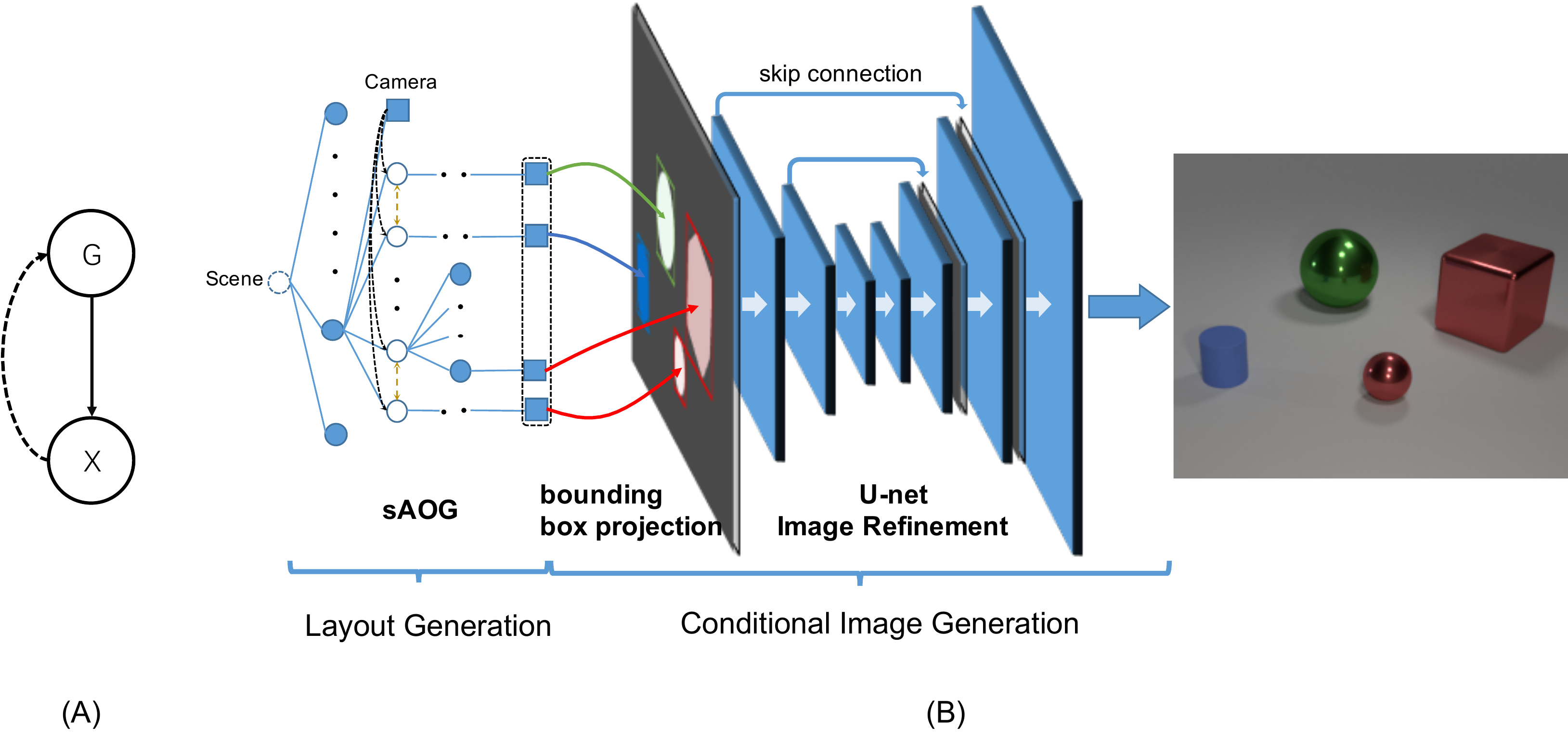}
	\caption{A) The probabilistic graphical model of our method. The solid line denotes the generative approach, and dash line denotes inference approach. In this model, $S$ is observed in training data, but need to be inferred for test data. B) An overview of the proposed method. A grammar model is firstly applied according to its and-node and or-node to generate a valid layout, and then map its bounding to image space. Then a refinement network is deployed to transfer the coarse image into a realistic image.}
	\label{fig:framework}
\end{figure*}

Our goal is to develop a deep structured generative model, where the structure is modeled by a grammar model, i.e. stochastic and-or graph(sAOG), and the generative model is deployed using a neural network. With bounding box and 3D location annotated, sAOG learns the intrinsic structure of a scene represented by a graph, and then converts the structure into image space by projecting objects as bounding boxes. Then a refinement model transfers the coarse bounding box map into a realistic image.

The training  dataset is denoted as $\mathcal{D} = \{( x_i, g_i)\}_{i=1}^N$ and structures in it as $D_g = \{g_i\}_{i=1}^N$, where $x_i$ notes raw image and $g_i$ denotes the structure. 
In this paper, $g_i = (O_i, E_i)$, $O_i = \{o_j^i\}_{j=1}^{n_i}, E_i = \{e_k^i\}_{k=1}^{m_i}$ where $o_j^i$ is the $j$-th attributed object in $i$-th image and $e_k^i$ is the $k$-th relation in $i$-th image. The attribute of an object includes shape, material, color, size, 3D location, and rotation. In the following, we use capital letter $X, G$ for random variables, and lower case $x,g$ for corresponding samples; $O,E$ respectively for the set of objects and relations in a scene, and $o,e$ for an object and a relation.

The graphical model of our method is illustrated in Fig.~\ref{fig:framework}. Accordingly, our proposed method is a latent variable model and the joint distribution $P(X,G)$ can be parameterized as:
\begin{eqnarray}
P(X, G) = P_G(G)P_{X|G}(X|G),
\end{eqnarray}
where $X$ represents the image and $G$ represents its structure. $P_G$ is the distribution over structures and $P_{X|G}$ is a conditional distribution of images. Compared to previous methods where the latent variables are generally a vector of random noise, the latent variable $G$ is defined by a stochastic and-or graph, and a sample is a parse graph $g$. The conditional distribution $P_{X|G}(X|G) = P_{X|M}(X|m(G))$, where $m(\cdot)$ maps structure $G$ into image space by projecting objects into 2-D bounding boxes from camera's view , and $P_{X|M}$ refines the bounding box map into a realistic image using a image-to-image model.

In the following, we first describe the detail of image grammar, and then the refinement network is illustrated.

\subsection{Grammar Model}
\begin{figure}
	\centering
	\includegraphics[width=0.45\textwidth]{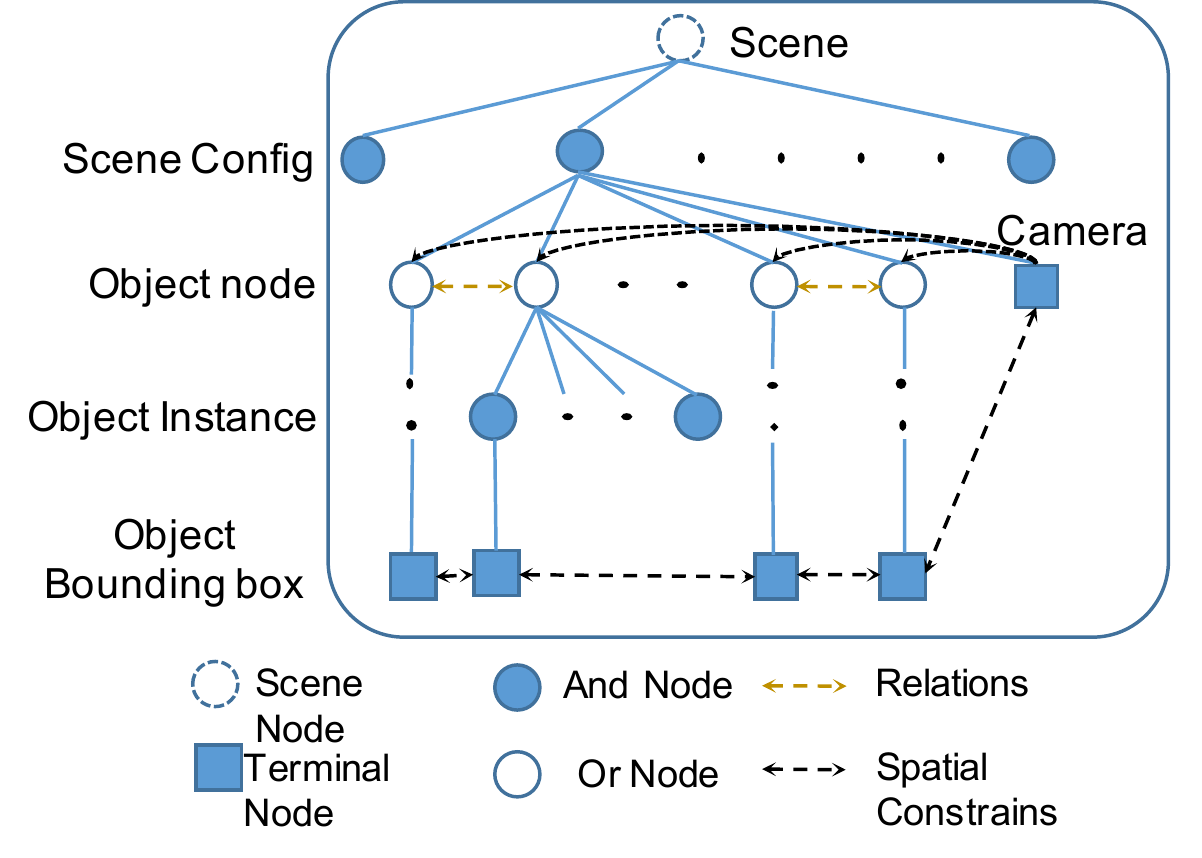}
	\caption{Structure of sAOG. It starts with a scene node, which randomly pick a scene configuration from its child nodes. A configuration node is an and-node composing $n_s$ object nodes and spatial relations and constrains are introduced between them. Each object node choose a instance independently, and the instances project their bounding box to a canvas to form a coarse image conditioned on the camera location.}
	\label{fig:saog}
\end{figure}
In this paper, the structure of our grammar model is illustrated in Fig.~\ref{fig:saog}, and the scene structure $g$ is represented as a parse graph. 
In a parse graph, the scene node $S$ first chooses a configure of a scene . 
In CLEVR dataset, a scene configure represents the number of objects in a scene. 
The configure is an and-node composing $n_s$ object nodes $\{ o_i \}_{i=1}^{n_s}$ (for convenience, we denote the node for the i-th object as $o_i$ ) and a camera node $c$. An object node is attributed with size, 3D locations and orientation. We denote all objects in a parse graph $g$ as $O_g = \{ o_i \}_{i=1}^{n_s}$. 
Relations among objects are introduced in this level, where the relations include front and right. Spatial constrains regularize the objects' 3D locations with certain relations. All relations in a parse graph $g$ is denoted as $E_g$, so each $e \in E_g$ represents a relation between certain two nodes.
The camera node is introduced to ensure all objects lie in the view of camera which means the location of each object should stay in certain range. 
Besides, we add a regularization on the height of location, guaranteeing objects are on the ground.
Then each object node selects a instance denoting its visual attributes such as color, shape, etc. Then the corresponding bounding box and mask are generated condition on the object location, visual attributes and camera's location.
In this paper, the camera's location is fixed for data, and this framework provides the flexibility to adjust the view point of certain scene.

Formally, the probability of a parse graph $g$ from the sAOG is decomposed as follows:
\begin{eqnarray}
P_G(g) = P_S(Ch_S)\prod_{o  \in O_g} P_{o}(Ch_{o}) \times P_\mathcal{E}(g),
\end{eqnarray}
where $Ch_{v}$ denotes the children of a non-terminal node v, $P_S$ is the probability of scene configuration, and $P_{o}$ is the probability of object instance condition on certain object node $o$. $P_\mathcal{E}(g)$ is the energy based distribution defined on the relations $E_g$ and other constrains. Specifically, the energy based distribution is defined as $P_\mathcal{E}(g) = \frac{1}{Z} exp(-\mathcal{E} (g))$ where $Z$ is the normalizing constnat. The energy function $\mathcal{E}$ is defined as follows:
\begin{eqnarray}
\mathcal{E}(g) = \lambda_d\sum_{\mathclap{e\in E_g}} \mathcal{E}_d(e) + \lambda_c\sum_{\mathclap{o\in O_g}} \mathcal{E}_c(o) + \lambda_h\sum_{\mathclap{o\in O_g}} \mathcal{E}_h(o).
	\label{eq:eng}
\end{eqnarray}

In Eq.\eqref{eq:eng}, (i) We define the set of relation types $\boldsymbol{R} = \{ \text{front}, \text{right}\}$, and $\mathcal{E}_d(e) = \max(\boldsymbol{n}_e^T \boldsymbol{r}_e, 0)$, where $\boldsymbol{n}_e$ is the direction of corresponding relation,  and $\boldsymbol{r}_e$ is the relative location between a pair of object in the relation. For example,  if $e$ means that object $o_1$ is on the right of $o_2$, then  $\boldsymbol{n}_e = (1,0,0)$ and $\boldsymbol{r}_e = \boldsymbol{r}(o_{e_1}) -\boldsymbol{r}(o_{e_2})  $, where  $\boldsymbol{r}(\cdot)$ is the location of an object. 
(ii) $\mathcal{E}_c$ is the energy for constrains of camera, and it's formulated as $\mathcal{E}_c(o) = -\log \tilde{P}(o,c)$, there $\tilde{P}$ is the smoothed empirical distribution of objects' location in camera $c$'s view. 
(iii) $\mathcal{E}_h$ is the energy to constraint object $o$'s bottom close to the ground. We define it as $\mathcal{E}_h(o) = |h_o|$ where $h_o$ is the height of object $o$'s bottom and the ground is zero in height.
In sum, we denote $\lambda = [\lambda_d, \lambda_c, \lambda_h]$ as the parameters that should be learned from data.

\subsection{Inference and Learning for Grammar}

To infer structure from an image, we apply an objection detection model to infer the 3D locations and other the attributes of objects . 
Given the attributed objects $O$, the relations $E$ can be inferred using maximum a posterior (MAP) as follows:
\begin{eqnarray}
E^* = \arg\max P_G(E|O),
\end{eqnarray}
where $P(E|O)$ is the posterior distribution of relations given node instance, and $E$ is the set of relations in the parse graph. MCMC method is applied to infer the MAP. Thus, MAP estimation of a parse graph is $g^*(O) =  (O, E^*)$ .

Specifically, a Faster-RCNN is used in our model to detect the bounding box and category for each object. After detection, the visual feature of each object, and its bounding box coordinates, are fed into a deep neural network to regress the 3D locations.
Relations can be inferred with MAP condition on the 3D locations with an MCMC method.

During training, maximum likelihood estimation (MLE) is applied to learn grammar on dataset $\mathcal{D}_g = \{g_i\}_{i=1}^N $ as follows:
\begin{eqnarray}
\max_{P_S, P_{o}, \lambda} \sum_{g\in \mathcal{D}_g} (\log P_G(g)) .
\end{eqnarray}
For each distribution of tree branch choosing $P_S$ and $P_o$, optimal distribution is simply the empirical distribution over the training data. For the weight parameters $\lambda$, we use contrastive divergence to optimize the parameters. To be specific, the update rule is illustrated as:
\begin{align}
\lambda_u^{t+1} & = \lambda_u^{t}-\alpha \left(
\frac{1}{|\mathcal{S}_t|} \sum_{O\in \mathcal{S}_t} \frac{\partial \mathcal{E}(g)}{\partial \lambda_u} 
-\frac{1}{|\mathcal{D}|} \sum_{g \in \mathcal{D_g}} \frac{\partial \mathcal{E}(g)}{\partial \lambda_u} 
\right)  \label{CD} \\
&= \lambda_u^{t}+\alpha \left(
\frac{1}{|\mathcal{S}_t|} \sum_{g\in \mathcal{S}_t} \mathcal{E}_{u}(g)
-\frac{1}{|\mathcal{D}|} \sum_{g \in \mathcal{D_g}}  \mathcal{E}_{u}(g)
\right), \label{ourCD}
\end{align}
where the subscript $u$ is either $c$, $e$ or $g$, and $\mathcal{S}_t$ is a set of parse graph we sampled from the grammar at $(t+1)$-th iteration.

\subsection{Conditional Image Generation Models}
\begin{figure*}[hbt!]
\centering
\includegraphics[width=\textwidth]{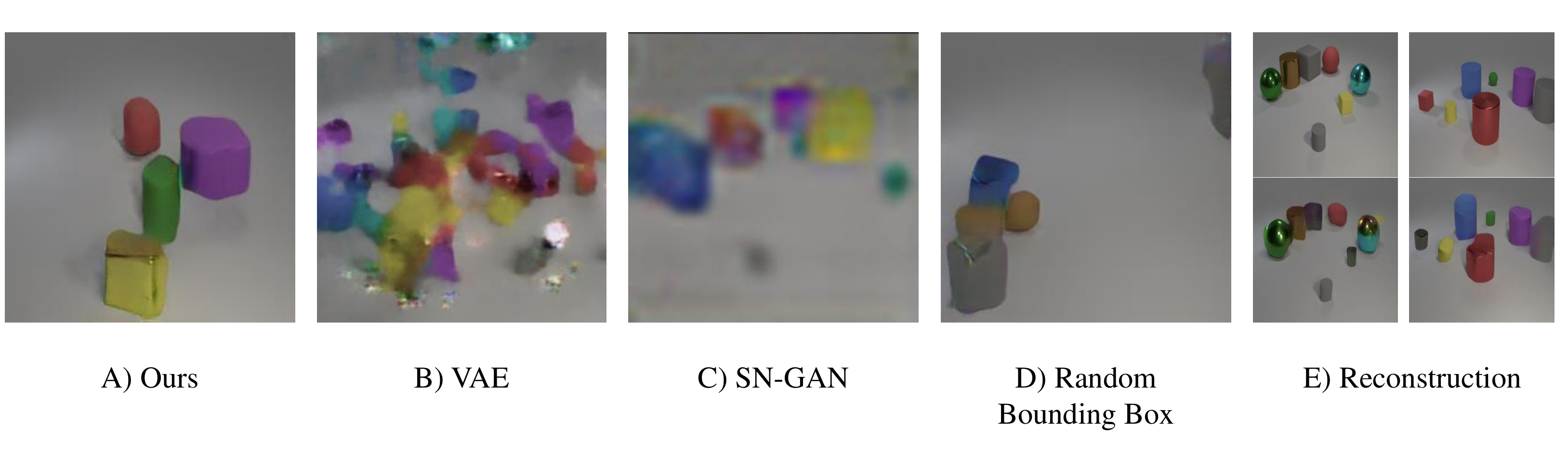}
\caption{ B) and C) is generated by  by VAE and SN-GAN.
A) is generated by our method conditioned on bounding box map of layouts from sAOG,
and D) by our refining network conditioned on randomly sampled bounding box.  In E), we demonstrate the reconstruct result of our proposed method. The upper line is the original image and the bottom line is the reconstructed images.}
\label{fig:exp_comp}
\end{figure*}

To implicitly model the conditional distribution $P_{X|G}$, we need to generate images conditioned on scene structures. This task is accomplished in two steps: bounding box projection and image refinement. In the first step, we manage to represent the information in scene structure $g$ as an image, by projecting the 3-D bounding box of each object in $g$ to the camera view, yielding an informative instance map. In this map, an object is represented as a 2-D bounding box, in which each pixel is a vector composed by a label embedding concatenated with a presentation for the orientation of the object. 

Thus, in the second step, it is natural to apply an image-to-image network to refine the instance map to a realistic image. In our method, we apply a modified pix2pix~\cite{isola2017image} to do this. Pix2pix model is composed by a U-net generator and a convolutional discriminator. The objective function contains two terms: a GAN loss that pushes the generated to the data distribution, and a L1 loss between fake images with real images which improves precision of image translation. 

In a U-net, a number of convolutional blocks encode an image into small size multi-channel feature map, and then equal number of deconvolutional blocks to recover an target image from the feature map. To conserve detailed information lost during convolution, in original U-nets, there are skip connections from each convolutional block to the  deconvolutional block on the same level. In our method, the refinement network should not only generate the appearance of objects, but also transform a bounding box to a proper shape of object, which is more challenging. To achieve the latter aim, we removed some skip connections on high levels of the U-net, so as to output more flexible object shapes.

\section{Experiment Result}
In this section, we evaluate our method with the widely used CLEVR dataset~\cite{johnson2017clevr} on both image generation and inference task. We first give a brief introduction of our experimental setup, then demonstrate the generated images from the whole framework, as well as the images reconstructed from the structure we inferred for a real image . With the grammar model embedded, several image editing operations can be applied in image generation, and it will be showed in the following section.

\begin{figure*}[hbt]
\includegraphics[width=\textwidth]{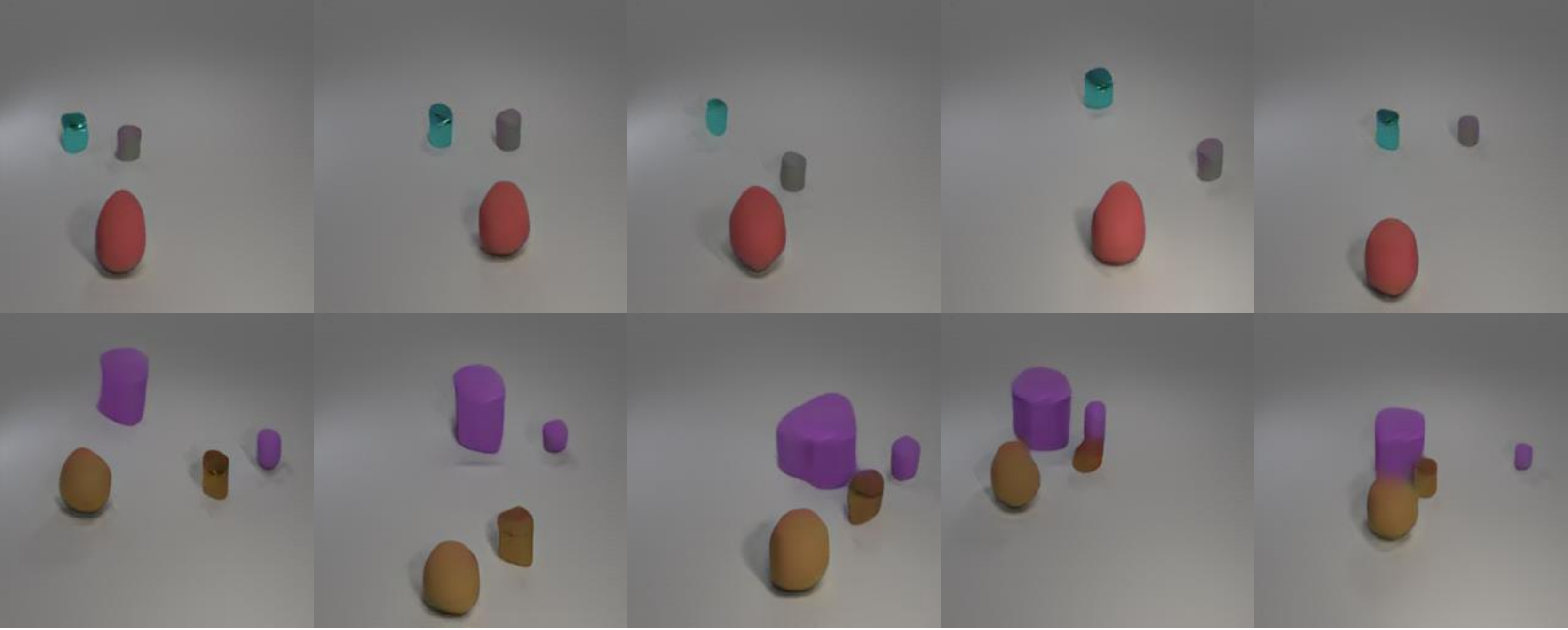}
\caption{ Images generated conditionally on spatial relations. Images in the same row are generated with same relations and attributes. The spatial relations are successfully preserved in generated images.}
\label{fig:condgen}
\end{figure*}

\subsection{Experimental Setup}
The CLEVR dataset consists of 70000 training images with resolution $480\times 320$, each consists of several simple objects such as cube, sphere and cylinder with 8 colors and 2 materials, totally 48 kinds of object labels. The 3D locations and spatial relations are given according to the location of the camera which is shared over the whole dataset. We use the provided attributes and 3D relations to build the sAOG, and then feed the projected bounding box images to refinement network for the realistic images.

A faster-rcnn with resnet-52 is used in the recognition model, and we use a two-layer MLP with 64 hidden neuron to regress the 3D locations of each object. The mean-square-error of the 3D location regression is $0.014$ which is quiet accurate and acceptable.

We use a U-net ~\cite{ronneberger2015u} structure as  the generative network, which has 8 convolution blocks and 8 deconvolution blocks. To allow shape transformation from bounding box to objects, there are no skip connection on the highest two levels of blocks. The U-net inputs an 9 channel bounding box instance map, where 3 channels are label embeddings represents 48 types of objects, and 6 channels mean the one-hot  vector of rotations discretized as $0^\circ\sim 15^\circ, 15^\circ\sim 30^\circ ... 75^\circ\sim 90 ^\circ$. The discriminator in this paper  is a 6-layer convolution network. 
Images are first resize to $256\times 256$, and then fed into the refinement network. After refined to a realistic image, we resize it back to $480\times 320$  which is the final output of our method.

\begin{figure*}[hbt]
\includegraphics[width=\textwidth]{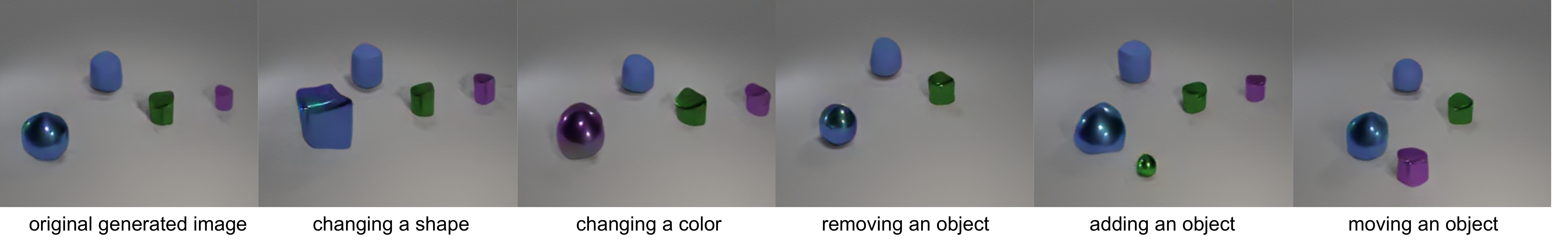}
\caption{Our method can accomplish fully generation, that is, sampling layouts and relations from sAOG and translate them into images. In this framework, user can easily editing the image on the scene structure, including manipulating the label or location of an object, moving or adding an object. }
\label{fig:edit}
\end{figure*}

\subsection{Image Generation}

In Fig.~\ref{fig:exp_comp}, we show the generated images from our proposed method, and VAE, SN-GAN. As we can see, the quality of images generated by our proposed method, surpass the quality of VAE and SN-GANs. Besides, we also demonstrate the generated images by random sampling a bounding box from the image space, and then transfer it as a realistic images. 

The sAOG gives an effective guidance of image generations for the refinement network, and the result demonstrates effectiveness of our proposed method. The images generated by VAE, especially,  is totally chaotic, and there's no reasonable structures of the generated images. The images generated by SN-GAN suffers from the same problem. For the randomly sampled bounding box, the generated image does not give a reasonable layout because of the collision of objects in 3D locations.

With an inference network, we can successfully compress an image into a configuration, and recover it using the refinement network. In Fig.\ref{fig:exp_comp}, column (E) gives examples of origin images and reconstructed images. As illustrated, the sAOG successfully recover the 3D layout and relations condition on a realistic images, and recover the semantic information. With successful inference on the spatial and semantic structures, we can compress an image with resolution $480\times 320$ using only 1KB.

\subsection{Conditional Generation}

With the grammar model embedded, we can generate images condition on certain structures, such as fix the relations between objects. The result of generated images is illustrated in Fig.~\ref{fig:condgen}. We generate a set of images with the relations and object attributes fixed. As we can see, the grammar model can successfully generate valid configurations according to the relation constrains and images with structure information can be generated conditionally. Besides conditional generation, the grammar model also enables us to do a set of operations, such as changing shape and color, removing or adding an object, and moving an object.  We demonstrate our results in Fig.\ref{fig:edit}. As we can see, this set of operation can be supported flexibly.

\section{Conclusion}
To summarize, we propose a novel framework that embeds grammar model into deep generative models. Our method successfully leverages the advance of sAOG and DGMs for complicated image generation. Experimental results demonstrate that sAOG can model the intrinsic structure of complicated relations between objects in an image, and can direct the refinement network to produce realistic images in a full generative approach. With a recognition model, structures can also be inferred from an image and multiple operations can be achieved with the highly interpretable and editable structures.
	
\section*{Acknowledgement}
This work was supported by NSFC Projects (Nos. 61620106010, 61621136008, 61332007), Beijing NSF Project (No. L172037), Tiangong Institute for Intelligent Computing, NVIDIA NVAIL Program, Siemens and Intel.

\bibliography{bibtex}
\bibliographystyle{icml2018}

\end{document}